\pdfoutput=1

\documentclass[11pt]{article}

\usepackage{emnlp2021}

\usepackage{times}
\usepackage{latexsym}
\usepackage{spverbatim}
\usepackage[T1]{fontenc}

\usepackage[utf8]{inputenc}

\usepackage{microtype}
\usepackage{times}
\usepackage{latexsym}
\usepackage{times}
\usepackage{latexsym}
\usepackage{hyperref}
\usepackage{url}
\usepackage{times}
\usepackage{latexsym}
\usepackage{times}
\usepackage{epsfig}
\usepackage{graphicx}
\usepackage{amssymb}
\usepackage{color}
\usepackage{amssymb}
\usepackage{fancyhdr}
\usepackage{listings}
\usepackage{multirow}
\usepackage{graphicx}
\usepackage{verbatim}
\usepackage{bbm}
\usepackage{color}
\usepackage{mathtools}
\usepackage{pifont}
\usepackage{makecell}
\usepackage{multirow}
\usepackage{booktabs}
\usepackage{float}
\usepackage{amsmath,amsfonts,amsthm,bm}
\usepackage{enumitem}
\usepackage{url}
\usepackage{pifont}
\usepackage{xcolor}
\usepackage{sidecap}
\usepackage{booktabs,siunitx} 
\usepackage{tikz}
\usepackage{array}
\usepackage{algorithmic}
\usepackage{algorithm}
\usepackage{subfig}
\usepackage{amsmath}

\DeclareMathOperator*{\argmax}{argmax}
\title{Towards Zero-Label Language Learning}

\author{Zirui Wang \qquad Adams Wei Yu \qquad  Orhan Firat \qquad Yuan Cao \\
Google AI \\
\texttt{\{ziruiw,adamsyuwei,orhanf,yuancao\}@google.com}}

\begin{document}
\maketitle
\begin{abstract}
This paper explores zero-label learning in Natural Language Processing (NLP),
whereby no human-annotated data is used anywhere during training and models are trained purely on synthetic data.
At the core of our framework is a novel approach for better leveraging the powerful pretrained language models.
Specifically, inspired by the recent success of few-shot inference on GPT-3,
we present a training data creation procedure named Unsupervised Data Generation (UDG),
which leverages few-shot prompts to synthesize high-quality training data without real human annotations.
Our method enables zero-label learning as we train task-specific models solely on the synthetic data, yet we achieve better or comparable results from strong baseline models trained on human-labeled data.
Furthermore, when mixed with labeled data, our approach serves as a highly effective data augmentation procedure,
achieving new state-of-the-art results on the SuperGLUE benchmark\footnote{Notably, our method is also the first to surpass human performance as of Dec 20, 2020.}.
\end{abstract}

\section{Introduction}

It is well-known that deep learning models are data-hungry. In natural language processing,
language model pre-training has become a successful transfer learning approach to effectively reduce the requirement for task-specific labeled data \cite{devlin2018bert,liu2019roberta,yang2019xlnet,radford2019language, raffel2019exploring,brown2020language}.
Via training on unsupervised large-scale text corpus, bi-directional language models such as BERT and XLNet are able to learn contextualized text representations that can then be fine-tuned on downstream tasks with small training data sizes, which have pushed the state of the art on a variety of natural language understanding benchmarks.

\begin{table}[t]
\begin{small}
\begin{center}
\begin{tabular}{l c c }
\toprule
\bf Model & \bf Setting & \bf SuperGLUE Avg. \\
\Xhline{\arrayrulewidth}
Human & & 89.8 \\
\Xhline{\arrayrulewidth}
Previous SOTA & \multirow{2}{*}{Supervised} & 89.3 \\
T5+UDG & & \bf 90.4 \\
\Xhline{\arrayrulewidth}
GPT3 & Few-Shot & 71.8 \\
\Xhline{\arrayrulewidth}
UDG & Unsupervised & \bf 78.1 \\
\bottomrule
\end{tabular}
\end{center}
\end{small}
\caption[caption]{SuperGLUE summary.}
\vskip -0.1in
\label{tab:superglue_summary}
\end{table}

\begin{figure*}[t]
    \centering
    \includegraphics[width=\textwidth]{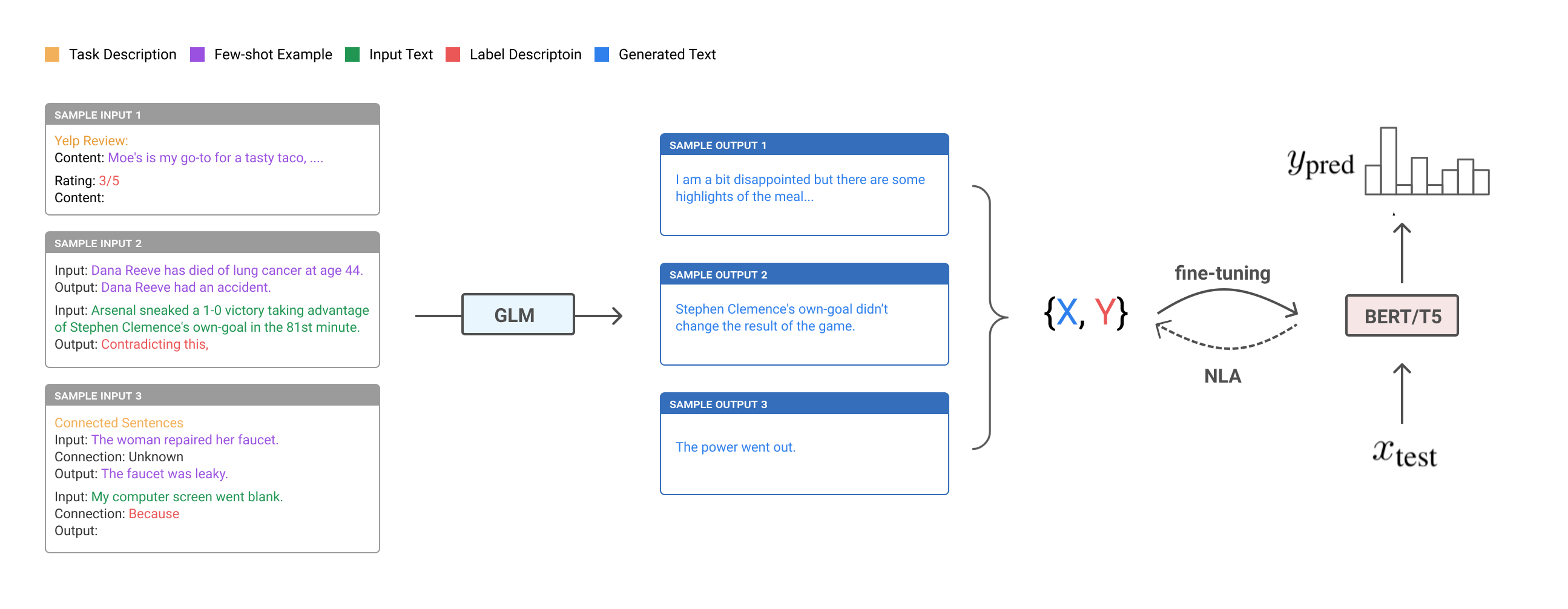}
    \caption{Illustration of the UDG framework.}
    \label{fig:method}
\vskip -0.1in
\end{figure*}

More recently, gigantic language models (GLM) such as GPT3 \cite{brown2020language} have been shown to be effective \textbf{few-shot learners}. As unsupervised training corpus and model size scaling up,
the model is able to generate answers for an unseen NLP task with few-shot inference,
based on a manually crafted input prompt consist of a task description and a few examples.
Despite no fine-tuning is involved,
the language model performs competitively against fine-tuned baselines on a wide range of tasks,
whose success suggests a new paradigm of transfer learning in NLP. Yet the gaps between few-shot inference and state-of-the-art fine-tuned methods are still large on many tasks (for example 17.5 below prior state-of-the-art on SuperGLUE as shown in Table \ref{tab:superglue_summary}),
urging for exploration of applications of giant language models beyond few-shot inference.

Inspired by the few-shot capability of GPT3, we shift our focus towards utilizing GLMs for example creation instead of direct inference,
and find that language models are also excellent \textbf{few-shot generators}.
Similar to the few-shot inference paradigm,
we query the model with a prompt with a few examples and a description of the desired label,
and the model generates examples aligned with the label while resembling the given samples.
Interestingly, we find no supervision is required for high-quality data creation and thus we only need to use unlabeled examples in our prompts. The dataset created by the model can then used to fine-tune any off-the-shelf model. This approach can therefore be treated as a \textit{zero-label} learning procedure, in which no human label is required throughout the whole process. It differs from the unsupervised learning procedure in that the downstream models still need to be trained with \textit{synthetic data}, however the training example creation requires no human labor.

Following this procedure, we are able to establish a system trained using unlabeled training data only, and thus we refer to it as \textbf{Unsupervised Data Generation} (\textbf{UDG}).
Experiments show that our unsupervised system performs competitively with strong supervised baselines and achieves new state-of-the-art few-shot learning results on text classification and the SuperGLUE language understanding benchmarks.
The synthesized data can further be used for data augmentation purpose. When combined with existing labeled data we are able to achieve the first super-human SuperGLUE scores.
These results suggest that few-shot training data creation is a promising alternative to few-shot inference with powerful language models.

\section{Related Work}

Data augmentation has traditionally been a popular technique for NLP model quality improvement, especially in low-resource regimes \cite{yu2018qanet,Wei19EDA}
While traditionally simple heuristics like token-level modification has been applied to diversify training samples, more recently generative data augmentation has gained popularity due to the progress made in language modeling \cite{anabytavor2019data,papanikolaou2020dare,juuti-etal-2020-little,lee2021neural,kumar2021data}.
However, they often require labeled examples to finetune generative models and heavy postprocessing for data cleaning.
On the other hand, our method generates data in a fully unsupervised manner without finetuning the language model,
showcasing a new zero-label learning paradigm.



Our approach is also closely related to knowledge retrieval from large language models. These models are known to be good at memorizing facts from training data and capable of performing as open knowledge bases
\cite{petroni2019language,wang2020language,roberts-etal-2020-much,carlini2021extracting}. The high quality of training examples created by our approach is to a large part guaranteed by the model's strong knowledge retrieval ability, which reduces the chance of erratic hallucinations irrelevant to the provided labels.

\begin{table*}[t]
\begin{center}
\resizebox{\textwidth}{!}{%
\begin{tabular}{l c c c c c c c c c}
\toprule
    & & \bf IMDb & \bf Yelp-2 & \bf Yelp-5 & \bf Amazon-2 & \bf Amazon-5 & \bf DBpedia & \bf Avg. \\
\toprule
XLNet & \multirow{2}{*}{Supervised} & \underline{96.80} & \underline{98.63} & \underline{72.95} & \underline{97.89} & \underline{68.33} & \underline{99.40} & \underline{89.00} \\
$\text{BERT}_{\text{LARGE}}$ & & 95.49 & 98.11 & 70.68 & 97.37 & 65.83 & 99.36 & 87.81 \\
\Xhline{\arrayrulewidth}
UDA & \multirow{2}{*}{Few-Shot} & 95.80 & 97.95 & 67.92 & 96.50 & 62.88 & 98.91 & 86.66\\
Few-shot Inf. & & 90.38 & 88.79 & 48.75 & 92.63 & 44.21 & 82.46 & 74.54 \\
\Xhline{\arrayrulewidth}
UDG & \multirow{2}{*}{Unsupervised} & 95.95 & 98.22 & 69.05 & 97.02 & 64.54 & 96.47 & 86.88\\
\enspace + NLA & & \bf 96.29 & \bf 98.38 & \bf 69.31 & \bf 97.24 & \bf 64.88 & \bf 99.21 & \bf 87.55 \\
\bottomrule
\end{tabular}
}
\end{center}
\caption[caption]{\textbf{Comparison of methods on text classification datasets (Accuracy)}. Results for XLNet are obtained from \cite{yang2019xlnet} while results for $\text{BERT}_{\text{LARGE}}$ and UDA are from \cite{xie2019unsupervised}. The best result for semi-supervised/few-shot setup is \textbf{bolded} while \underline{underline} signifies the overall best.}
\vskip -0.1in
\label{tab:text_classification}
\end{table*}

\section{Method}

\subsection{Background: Few-shot Inference}

Given a set of labeled data $\mathcal{L} = \{(x^i, y^i)\}_{i=1}^{n}$ for a specific downstream task,
the most common approach in recent years has been \textbf{fine-tuning} that updates the weights of a pre-trained model according to $\mathcal{L}$ \cite{devlin2018bert, yang2019xlnet,raffel2019exploring}.
While obtaining state-of-the-art performance on a wide range of tasks, fine-tuning requires extra update steps and non-trivial amounts of labeled data in the target task.
On the other hand, \textbf{few-shot inference} is a more resource-efficient paradigm exhibited in the latest gigantic language models such as GPT3 \cite{radford2019language,brown2020language}.
The idea is to utilize the language model to infer the correct label based on the task description and a few sample input-label pairs.
In particular, the input to the model $M$ is a handcrafted ordered prompt consisted of a task description $T$, a small set of K examples $\mathcal{L_{\text{few}}} = \{(x^i, y^i)\}_{i=1}^{K} \subseteq \mathcal{L}$, and the query example $x_{q}$,
and the model is expected to infer the correct label $y_q$ as the most probable next text sequence to the input prompt:
\begin{equation}
    y_{q} = \argmax_{y} P_{M}(y| [T, \mathcal{L_{\text{few}}}, x_{q}] ).
\end{equation}
Since taking the argmax is intractable, $y_{q}$ is usually obtained through greedy decoding or beam search.
Using much less task-specific data and no gradient update, few-shot inference can obtain performance comparable to fine-tuning methods (e.g. GPT3 performs similarly to fine-tuned BERT on SuperGLUE in Table \ref{tab:superglue}).
In its extreme format, giant language models can also perform one-shot (K=1) or even zero-shot (K=0) inference.

\subsection{Unsupervised Data Generation}

Despite these interesting findings, 
few-shot inference using giant language models still underperforms state-of-the-art fine-tuned models on many tasks.
In Table \ref{tab:superglue}, for instance, T5 largely outperforms GPT3 (89.3 vs 71.8) despite being much smaller in model sizes (11B vs 175B).
One potential limitation is that a language model is never explicitly trained to directly conduct inference.
Instead, it is trained as a text generator on unsupervised web corpus where inputs ($X$) and labels ($Y$) happen to coexist.
Consequently, the few-shot inference method finds the proper prompt that `forces' the model to generate next text sequence $X_{\text{next}}$ which happens to be the label Y.
However, this could be suboptimal since the labels often emerge prior to the inputs in real-world web documents.
For example, in sentiment classification of IMDb movie reviews \cite{maas2011learning}, the actual review contexts appear after their corresponding rating scores.
Therefore, few-shot inference can force the language model to generate on text distributions that are inconsistent with its training data.

To this end, we propose to utilize language models to perform \textbf{few-shot generation}.
Instead of generating and predicting the label Y, we let the model to generate the input X instead,
decoupling generation from prediction.
We aim to formulate the input prompts that are more likely to naturally exist in the training corpus.
Specifically, the model is queried to generate $x_g$ corresponding to a pseudo label $\hat{y}_{g}$
with a prompt consisted of a small set of K \textit{unlabeled} examples $\mathcal{U} = \{x^i\}_{i=1}^{K}$ and a description of the desired label:
\begin{equation}
    x_g \sim P_M(x| [T, \mathcal{U}, \text{Des}(\hat{y}_{g})] ),
\end{equation}
where $\text{Des}(\cdot)$ is a task-specific transformation function that maps a label class to natural language descriptions,
as illustrated in Figure \ref{fig:method}.
Different from few-shot inference, our method only requires unsupervised few-shot examples, a \textit{zero-label learning} setting.
In addition, we use top-k sampling instead of search-based decoding to sample text from the language model.
This allows us to generate a synthetic labeled dataset $\mathcal{L_{\text{syn}}} = \{(x_g^i, \hat{y}_g^i)\}_{i=1}^{n_s}$ with controllable size $n_s$.
We then train task-specific models utilizing this synthetic dataset,
either as standalone training data or additional auxiliary data.
Unlike existing synthetic data generation systems,
our method requires no fine-tuning step of the generative model and uses unsupervised data only,
and therefore we refer to it as \textit{Unsupervised} Data Generation to emphasize its resource efficiency. We also hope to emphasize that it is not our intention to leverage the language model to perform generative tasks, but just to take advantage of it to synthesize ``labeled'' examples for downstream model training.

\section{Experiments}

\begin{table}[t]
\begin{center}
\begin{tabular}{l c c c c c}
\toprule
 & K=0 & K=1 & K=4 & K=32 \\
\Xhline{\arrayrulewidth}
\bf IMDb Acc. & 64.21  & 91.34 & 95.86 & 96.29 \\
\bf Yelp-2 Acc. & 67.34 & 90.27 & 98.22 & 98.38 \\
\bf Amz-5 Acc. & 47.35 & 58.79 & 62.14 & 64.88 \\
\bottomrule
\end{tabular}
\end{center}
\caption[caption]{Ablation of number of examples in each prompt.}
\vskip -0.1in
\label{tab:ablation_K}
\end{table}

\subsection{Unsupervised Text Classification}

We first apply the proposed UDG method on standard text classification tasks.

\noindent \textbf{Experimental Setups.} 
We use six popular text classification benchmark datasets \cite{maas2011learning,zhang2015character},
including IMDb, Yelp-2, Yelp-5, Amazon-2 and Amazon-5 sentiment classification and DBPedia topic classification.
We mainly follow the experimental settings in \citet{xie2019unsupervised} and use the corresponding unlabeled data for each task.
We apply similar preprocessing steps to clean noisy web texts and truncate the input to 512 subword tokens.
For each prompt, we sample $K=32$ unlabeled examples from the unlabeled data and fit as many examples as allowed by the length of the language model's context window (detailed templates shown in Figure \ref{fig:method} and Appendix \ref{sec:appendix_template}).
This process is then repeated $n_c = \frac{n_s}{\text{\# Class}}$ times for each label class,
where we set $n_c = 10\text{K}$ for sentiment classification tasks and 1000 for topic classification.
We then utilize the language model to generate one example for each prompt,
resulting in a synthetic labeled dataset of size $n_s$.
We use an in-house language model, which is a variant of the one in \cite{adiwardana2020towards} but trained with larger data.
We exploit top-k sampling with K=40 and temperature=1.0, and only apply basic post-processing to filter generated examples that are too short/long.

Once we obtain the generated synthetic dataset $\mathcal{L_{\text{syn}}}$,
it can be utilized as labeled training data for any task-specific training framework.
Here, we choose the state-of-the-art semi-supervised learning framework Unsupervised Data Augmentation (UDA) \cite{xie2019unsupervised} as the backbone.
We use $\text{BERT}_{\text{Large}}$ as our base model and follow the training protocol as described in the UDA paper to tune our hyper-parameters.
In our experiment, we find some generated examples are noisy adn thus we additionally implement a \textit{Noisy Label Annealing (NLA)} technique to filter these examples during the training process (See Appendix \ref{sec:appendix_nla} for details).

\begin{figure}
    \centering
    \includegraphics[width=0.6\columnwidth]{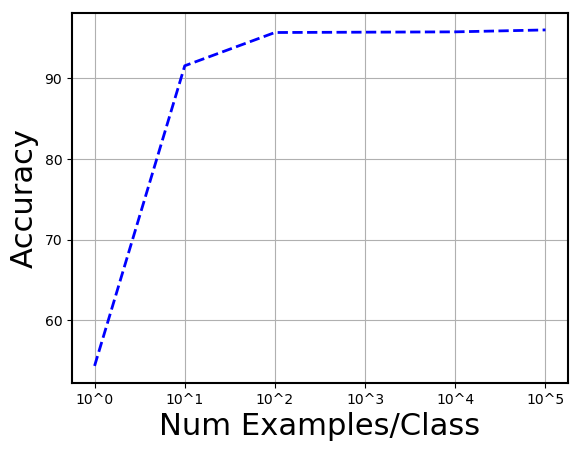}
    \caption{Ablation of number of examples generated per label class.}
    \label{fig:ablation_ns}
\vskip -0.1in
\end{figure}

\noindent \textbf{Results.} 
We compare models of trained using fully supervised, semi-supervised/few-shot and unsupervised settings in Table \ref{tab:text_classification}.
We first compare few-shot inference using our giant language model with fine-tuned methods.
Despite requiring no additional training costs, the few-shot inference paradigm performs significantly worse than supervised or even semi-supervised UDA, which utilizes similar amounts of labeled data.
The gap is more evident on multi-way classification tasks such as Yelp-5 or DBpedia, where the model is required to predict complex labels beyond simple answers such as `True/False'.
In contrast, the proposed few-shot generation paradigm obtains strong performance while using less supervision.
When combined with NLA, our UDG framework consistently outperforms UDA and few-shot inference on all six tasks, achieving new state-of-the-art few-shot learning results.
Besides, without using any label, our method outperforms fully supervised $\text{BERT}_{\text{LARGE}}$ on IMDb and Yelp-2 and is also competitive on other tasks. 
Since both UDA and our method rely on $\text{BERT}_{\text{LARGE}}$, we expect using XLNet may further boost our unsupervised performance, which we choose to leave for future work.

\begin{table*}[t]
\begin{center}
\resizebox{\textwidth}{!}{%
\begin{tabular}{l c c c c c c c c c c}
\toprule
& & \bf BoolQ & \bf CB & \bf COPA & \bf MultiRC & \bf ReCoRD & \bf RTE & \bf WiC & \bf WSC & \bf Avg. \\
\toprule
Human & & 89.0 & 95.8/98.9 & 100.0 & 81.8/51.9 & 91.7/91.3 & 93.6 & 80.0 & 100.0 & 89.8 \\
\Xhline{\arrayrulewidth}
$\text{BERT++}^{\text{a}}$ & \multirow{5}{*}{Sup.} & 79.0 & 84.8/90.4 & 73.8 & 70.0/24.1 & 72.0/71.3 & 71.7 & 69.6 & 64.4 & 71.5 \\
$\text{RoBERTa}^{\text{b}}$ & & 87.1 & 90.5/95.2 & 90.6 & 84.4/52.5 & 90.6/90.0 & 88.2 & 69.9 & 89.0 & 84.6 \\
$\text{T5}^{\text{c}}$  & & 91.2 & 93.9/96.8 & 94.8 & 88.1/63.3 & 94.1/93.4 & 92.5 & 76.9 & 93.8 & 89.3 \\
$\text{DeBERTa}^{\text{d}}$  & & 90.4 & 94.9/97.2 & 96.8 & \bf 88.2/63.7 & \bf 94.5/94.1 & \bf 93.2 & 76.4 & 95.9 & 89.9 \\ 
T5 + UDG  & & \bf 91.4 & \bf 95.8/97.6 & \bf 98.0 & 88.3/63.0 & 94.2/93.5 & 93.0 & \bf 77.9 & \bf 96.6 & \bf 90.4  \\
\Xhline{\arrayrulewidth}
$\text{GPT3}^{\text{e}}$  & \multirow{3}{*}{Few-Shot} &76.4 & 52.0/75.6 & \underline{92.0} & 75.4/30.5 & \underline{91.1/90.2} & 69.0 & 49.4 & 80.1 & 71.8 \\
$\text{iPET}^{\text{f}}$  & & \underline{81.2} & 79.9/88.8 & 90.8 & 74.1/31.7 & 85.9/85.4 & 70.8 & 49.3 & \underline{88.4} & 75.4 \\
$\text{ADAPET}^{\text{g}}$  & & 80.0 & 82.3/92.0 & 85.4 & 76.2/35.7 & 86.1/85.5 & 75.0 & 53.5 & 85.6 & 76.0 \\
\Xhline{\arrayrulewidth}
UDG & Unsup. & 81.0 & \underline{86.2/92.4} & 80.4 & \underline{81.1/47.1} & 82.8/81.8 & \underline{80.7} & \underline{67.5} & 79.5 & \underline{78.1} \\
\bottomrule
\end{tabular}
}
\end{center}
\caption[caption]{\textbf{Comparison of single-model methods on SuperGLUE test scores.} Results obtained from the official SuperGLUE leaderboard\protect\footnotemark. The best result for semi-supervised/few-shot setup is \underline{underlined} while \textbf{bold} signifies the overall best. Model references: $^{\text{a}}$\citet{devlin2018bert} $^{\text{b}}$\citet{liu2019roberta} $^{\text{c}}$\citet{raffel2019exploring} $^{\text{d}}$\citet{devlin2018bert} $^{\text{e}}$\citet{brown2020language} $^{\text{f}}$\citet{schick2020s} $^{\text{g}}$\citet{tam2021improving}}
\vskip -0.1in
\label{tab:superglue}
\end{table*}

\noindent \textbf{Analysis.} 
We first examine the effect of data noisiness on model performance.
As is the case for other data augmentation methods, few-shot generation using giant language models can produce examples that are inaccurate to the desired labels.
To reduce the negative impact of these noisy labels,
we utilize a simple NLA technique to filter out examples when the task-specific models disagree with the synthetic labels with high confidence levels.
As shown in Table \ref{tab:text_classification}, NLA robustly improves UDG performance on all tasks,
especially ones that are sensitive to noise such as DBpedia.

A crucial difference distinguishing our work from existing data generation methods is that we directly query the pretrained language model without any fine-tuning nor supervision.
To achieve this, the model needs to not only infer correct knowledge corresponding to the input pseudo label
but also generate text with similar styles of the sample unsupervised examples.
Thus, we compare the results when the language model uses different amounts of in-context examples in Table \ref{tab:ablation_K}.
The model fails to generate high-quality data when no sample is given, indicating the importance of few-shot generation.
On the other hand, including more unsupervised examples does improve the quality of synthetic dataset which leads to better performance.

Finally, we evaluate the impact of the synthetic data sizes in Figure \ref{fig:ablation_ns}.
Despite there is a diminishing return trend, we find the final performance to continuously improve with more generated data, showing that the language model can generate diverse examples.
In addition, one key benefit of our method is that we can sample as much data as needed with no additional cost or supervision.
This is particularly useful for tasks from low-resource domains with limited unsupervised data available.

\subsection{Unsupervised Language Understanding}

To evaluate the proposed framework in a more challenging and comprehensive setting,
we extend it to perform on complex language understanding tasks.

\noindent \textbf{Experimental Setups.} We use the SuperGLUE benchmark \cite{wang2019superglue} for general-purpose language understanding in English,
which consists of 8 natural language understanding tasks.
Tasks cover textual entailment (CB and RTE),
question answering (BoolQ, MultiRC and ReCoRD),
common sense reasoning (COPA),
word sense disambiguation (WiC),
and coreference resolution (WSC).
We mainly follow the same generation protocol as described in the previous sections,
with some minor changes in prompt templates and data post-processing steps for specific tasks.
As before, we use K=32 unlabeled examples and generate using the same language model.
For each task, we use all original labeled data as unsupervised examples for training data creation.

For the downstream model, we use T5 \cite{raffel2019exploring} for fine-tuning on the created data.
Different from the released T5 checkpoints that are pretrained on multi-task data,
we pretrain our own models on unsupervised Colossal Clean Crawled Corpus (C4) data only and thus the combined framework remains unsupervised.
For fair comparison with existing models, we pretrain and then fine-tune a T5-Large model using the created data set.
Following \citet{raffel2019exploring}, we use a fine-tuning batch size of 8 with 512 sequence length.

\footnotetext{\url{https://super.gluebenchmark.com/leaderboard}}

\noindent \textbf{Results.}
We compare models trained under different settings in Table \ref{tab:superglue}.
The GPT3 model \cite{brown2020language} using the few-shot inference method outperform BERT++ with less supervision and no fine-tuning.
However, despite containing much more model parameters, it performs worse than other fine-tuned fully supervised models and few-shot methods.
On the other hand, our unsupervised framework using few-shot generation outperforms all few-shot learning systems without using any label,
and thus it achieves new state-of-the-art results on this benchmark for methods that exploit little-to-no supervision.
In particular, our performance gains largely come from natural language entailment tasks (CB and RTE) as well as word sense disambiguation, where GPT3 performs similarly to random guessing.
This indicates that language models do contain language knowledge that few-shot inference fails to leverage.

\subsection{UDG as Data Augmentation}

In previous sections we only use the created examples as pseudo supervision to explore the limits of transfer learning using language models.
Nonetheless, the synthetic data can be also treated as augmented data and combined with existing labeled data.
To this end, we fine-tune the public T5-XXL checkpoint using both labeled data and generated data.
As shown in Table \ref{tab:superglue}, our method combines well with existing labeled data and brings substantial improvements.
This is particularly the case for tasks with small data sizes such as COPA and WSC.
Moreover, the combined model outperforms not only prior methods but also the human baselines for the first time on this important NLP benchmark,
setting a new milestone for natural language understanding with machine learning models.

\section{Conclusion}

In this paper, we propose a ``zero-label'' training procedure and show that language models are also few-shot example creators in that they can be used to generate high-quality synthetic data in a fully unsupervised manner. Through this, we demonstrate that NLP models can obtain strong results without any human annotated label. Our work illustrate a promising direction for future transfer learning research in NLP.

\bibliography{anthology,custom}
\bibliographystyle{acl_natbib}

\clearpage

\appendix

\section{Noisy Label Annealing}
\label{sec:appendix_nla}

Noisiness is a common issue for synthetic data generation.
To mitigate this issue, prior work [CITE] utilize extensive filtering methods to select clean generated examples.
While one key benefit of our method being high-quality synthetic data with minimal filtering,
we do find some regularization during finetuning to be helpful for better performance,
especially on tasks sensitive to noises.
In particular, we obverse that the generated examples of the language model may be misaligned with the desired label class.
Thus, we introduce a new training technique called Noisy Label Annealing (NLA),
which gradually filter out noisy training signals as training progresses. Intuitively, we remove a specific training example if our model disagrees with its label with high confidence.
Mathematically, at training step t, a given example $(x_g^i, \hat{y}_g^i)$ is considered noisy and removed,
if (1) the model's predicted probability $P(y|x_g^i)$ is higher than a threshold $\mu_t$,
and (2) the prediction $\overline{y}^i = \argmax_{y} P(y|x_g^i)$ differs from the synthetic label $\overline{y}^i \neq \hat{y}_g^i$.
We set the initial threshold $\mu_0$ to 0.9 and gradually anneal it to $\frac{1}{K}$ where $K$ is the number of classes.
Intuitively, the model is less accurate at the early stage of the finetuning process and thus we demand a very high confidence level to filter noises,
whereas we can safely decrease the ``bar'' as the model gets better trained.
We explore different final annealing values in Table \ref{tab:ablation_nla} and find a more aggressive strategy works often better.

\begin{table}[t]
\begin{center}
\resizebox{0.5\textwidth}{!}{%
\begin{tabular}{c c c c c c}
\toprule
None & 0.9$\rightarrow$0.8 & 0.9$\rightarrow$0.7 & 0.9$\rightarrow$0.6 & 0.9$\rightarrow$0.5 \\
\Xhline{\arrayrulewidth}
95.95 & 96.03 & 96.08 & 96.17 & 96.29 \\
\bottomrule
\end{tabular}
}
\end{center}
\caption[caption]{Comparison of different annealing thresholds on IMBd classification. We observe performance improves as we filter more aggresively.}
\vskip -0.1in
\label{tab:ablation_nla}
\end{table}

\section{Finetuning Details}
\label{sec:appendix}

For text classifications, we mainly follow the experimental setups in \cite{xie2019unsupervised}.
We truncate the input to 512 subwords using BERT's vocabulary, keeping the last tokens.
For the finetuning process, we search the learning rate in \{1e-5, 2e-5, 5e-5\} and batch size in \{32, 64, 128\}.
We also tune the number of epochs based on the size of generated data, ranging from 5 to 30.
As with \cite{xie2019unsupervised}, we also fine-tune the BERT model on in-domain unsupervised data prior to the final training stage.
For UDA hyperparameters, we tune the batch size and weight for both unsupervised and generated data, as well as different strategies of Training Signal Annealing (TSA).
Notice that TSA is orthogonal to our NLA technique and thus we can apply them at the same time.
Experiments are conducted on 32 v3 TPUs.

For tasks in SuperGLUE, we follow the pretraining and finetuning setups in \cite{raffel2019exploring}.
For the unsupervised setting, we pretrain our own T5-Large model using the exact same settings and hyperparameters as the original paper,
except that we do not utilize the original downstream tasks data for multi-task pretraining.
For all finetuning, we use a batch size of 8 and a sequence leangth of 512, and utilize the dev scores to pick the best checkpoint for submission to the test server for final evaluation.
For more details, we refer readers to the T5 paper \cite{raffel2019exploring}.

\clearpage
\onecolumn
\section{Prompts \& Generated Examples}
\label{sec:appendix_template}

\subsection{Text Classification}
\subsubsection{IMDb}
\textbf{Input Prompt:}
\begin{spverbatim}
Sample Movie Review: This is the most saccharine romance I ever sat through. The perfect film for an idle housewife in kerchief, housedress, and ostrich-trimmed high-heeled mules to watch in the afternoon, lying on the couch eating bonbons. In fact, bonbons play a prominent role in the movie.  The only reason I was able to watch to the end, is that I finally was able to gaze at Keanu Reeves' dreamy face in almost every scene. In most of his films, he moves too fast to get a good look. The only rapid action in this show is Giancarlo Giannini waving his hands with Latin emotionality - more Italian than Mexican, really.

Negative Movie Review:
\end{spverbatim}

\vspace{10px}

\noindent \textbf{Generated Output:}
\begin{spverbatim}
This remake of the 1966 French film by François Truffaut and Claude Chabrol is just plain awful. The only positive I can say about the film is that the casting was just about perfect. Otherwise, the movie has no redeeming qualities whatsoever. The plot is a mess and is filled with numerous inconsistencies and continuity flaws throughout. It is as if the director just made it up on the spot and made it up bad!
\end{spverbatim}

\subsubsection{Yelp-2}
\textbf{Input Prompt:}
\begin{spverbatim}
Sample Business Review: I had the most delicious burger ever! I also ordered the truffle fries, which were amazing! The sauce they served with the fries was also very good! Definitely come here!! There was great service and wonderful food!

Positive Business Review:
\end{spverbatim}
\vspace{10px}

\noindent \textbf{Generated Output:}
\begin{spverbatim}
We had our wedding at The Grand in October of 2011. The Grand and their staff are absolutely wonderful to work with. We had over 100 people at the wedding and the room was packed for the reception. We had a great time and the service, food, and atmosphere were fantastic. I can not say enough good things about working with and having our wedding at The Grand. We highly recommend The Grand for your wedding or special event.
\end{spverbatim}

\subsubsection{Yelp-5}
\textbf{Input Prompt:}
\begin{spverbatim}
Yelp Review
Content: I waited FOR 35 MINUTES tonight just for 2 brisket meals to go. They weren't even hot when I arrived home and I live less than 1 minute away. It was still a fine meal, saved by the amazing cornbread!! But, I have no intention of going back any time soon.

Yelp Review
Rating: 3/5
Content:
\end{spverbatim}
\vspace{10px}

\noindent \textbf{Generated Output:}
\begin{spverbatim}
I came here to eat after the ridiculous long lines died down a bit. I must say, I don't get the hype about this place and felt a bit disappointed. I suggested coming here with a group of friends. I was pretty excited to eat here and found that nothing impressed me and the price tag could not be justified. Nothing stood out for me that would make me want to come here again. I think the highlight of my meal was the ginseng chicken soup. All other items, I've had better at other buffets. I didn't think their selection was that impressive.
\end{spverbatim}

\subsubsection{Amazon-2}
\textbf{Input Prompt:}
\begin{spverbatim}
Sample Product Review
Title: Nice to have
Content: My dog loves this bed. I don't like to have my dog sleep on the floor. I know I spoiled my dog. I put a huge pillow on top of this bed to give her the extra comfort. My dog loves sleeping on something soft. Now I have a happy dog that sleeps comfortably every night. Money well spend.

Negative Product Review
Title:
\end{spverbatim}
\vspace{10px}

\noindent \textbf{Generated Output:}
\begin{spverbatim}
Not worth it
Content: I am so very disappointed. I bought this for my granddaughter for Christmas. I have a few concerns, but first and foremost, the box that the doll comes in says that it must be assembled by an adult. The instructions are very confusing. My mom and I put it together for her Christmas Eve. Then we realized that the doll has a small hole in the back of the head that is on the end of the seam. I don't know where the hole came from, but since she is a collectible, I can't return it.
\end{spverbatim}

\subsubsection{Amazon-5}
\textbf{Input Prompt:}
\begin{spverbatim}
Product Review
Title: Good Book
Content: Good book, easy to read, with a lot of suggestions about where read the exposure. Also if the book is more than 150 pages, there are a lot of big photos, everyone with a comment; instead of this I would have prefereed more photos with comment also if they would have been smaller. So, I read this book in too little time. My rate for this book is 3.5.


Product Review
Rating: 3/5
Title:
\end{spverbatim}
\vspace{10px}

\noindent \textbf{Generated Output:}
\begin{spverbatim}
An ok game
Content: This game is good, but I wouldn't buy this game unless you are a star wars fan. The game has lots of levels and it takes quite a while to beat. If you don't know anything about star wars, you would be lost on the story. There are lots of good things and bad things about this game.
\end{spverbatim}

\subsubsection{DBpedia}
\textbf{Input Prompt:}
\begin{spverbatim}
Wikipedia Article
Title: University of Georgia School of Law
Article:  The University of Georgia School of Law (also referred to as Georgia Law) is a graduate school of the University of Georgia.


Wikipedia Article
Category: Educational Institution
Title:
\end{spverbatim}
\vspace{10px}

\noindent \textbf{Generated Output:}
\begin{spverbatim}
Vanung University
Article: Vanung University (VNU) is a private university in Zhongli City Taoyuan County Taiwan. It is the first private university in Taiwan to be accredited by AACSB.
\end{spverbatim}

\subsection{SuperGLUE}
\subsubsection{BoolQ}
\textbf{Input Prompt:}
\begin{spverbatim}
Article: Russian Blue -- Russian Blues are plush short-haired, shimmering pale blue-gray cats with emerald green eyes. Guard hairs are distinctly silver-tipped giving the cat a silvery sheen or lustrous appearance. They have been used on a limited basis to create other breeds such as the Havana Brown or alter existing breeds such as the Nebelung. They are being used in Italy as a way to make Oriental Shorthairs healthier and more robust called RUS4OSH in FIFe.
Sample statement based on the article: do all russian blue cats have green eyes.

Article: Okinawa Churaumi Aquarium -- The Okinawa Churaumi Aquarium (Okinawa Churaumi Suizokukan) is located within the Ocean Expo Park in Okinawa, Japan. It welcomed its 20 millionth visitor on 30 March 2010 and is a member of the Japanese Association of Zoos and Aquariums (JAZA). It was the largest aquarium in the world until it was surpassed by the Georgia Aquarium in 2005. The aquarium has the theme of ``Encounter the Okinawan Sea''.
False statement based on the article:
\end{spverbatim}
\vspace{10px}

\noindent \textbf{Generated Output:}
\begin{spverbatim}
is okinawa churaumi aquarium the largest aquarium in the world.
\end{spverbatim}

\subsubsection{CB}
\textbf{Input Prompt:}
\begin{spverbatim}
Input: Jim waited. He waited a long time, and when the young doctor finally came out, it was almost dark. Jim could nonetheless tell by his anxious face that something was wrong.
Output: Something was wrong.

Input: She swallowed hard, unsure if she had the nerve to go ahead. The memory of the pain in Tara's eyes last night decided her. Did he really expect her to believe that Tara was only the housekeeper?
Output: Contradicting this, 
\end{spverbatim}
\vspace{10px}

\noindent \textbf{Generated Output:}
\begin{spverbatim}
Tara was only a temporary housekeeper.
\end{spverbatim}

\subsubsection{COPA}
\textbf{Input Prompt:}
\begin{spverbatim}
Input: My body cast a shadow over the grass.
Output: The sun was rising. 

Input: My computer screen went blank.
Connection: Because
Output:
\end{spverbatim}
\vspace{10px}

\noindent \textbf{Generated Output:}
\begin{spverbatim}
The power went out.
\end{spverbatim}

\subsubsection{MultiRC}
\textbf{Input Prompt:}
\begin{spverbatim}
Article: On a day it celebrated 50 years of giving free legal advice, Southern Arizona Legal Aid announced it has received two grants totaling more than 7 million ...
Question: Who will the Web site benefit?
Answer: Help bridge the "digital divide" that keeps the poor from using the Internet as a resource

Article: While this process moved along, diplomacy continued its rounds. Direct pressure on the Taliban had proved unsuccessful ...
Question:
\end{spverbatim}
\vspace{10px}

\noindent \textbf{Generated Output:}
\begin{spverbatim}
Which country did President Bill Clinton visit on March 25, 2000?
Correct Answer: Pakistan
\end{spverbatim}

\subsubsection{ReCoRD}
\textbf{Input Prompt:}
\begin{spverbatim}
Article: The harrowing stories of women and children locked up for so-called 'moral crimes' in Afghanistan's notorious female prison have been revealed after cameras were allowed inside. Mariam has been in Badam Bagh prison for three months after she shot a man who just raped her at gunpoint and then turned the weapon on herself - but she has yet to been charged. Nuria has eight months left to serve of her sentence for trying to divorce her husband. She gave birth in prison to her son and they share a cell together. Scroll down for video Nuria was jailed for trying to divorce her husband. Her son is one of 62 children living at Badam Bagh prison.
Keyword: Badam Bagh prison ### Relevant Fact: The women and children are jailed in Badam Bagh prison.

Article: China's space agency today distributed fresh new pictures of the Moon's surface from its lunar lander and rover. The Jade Rabbit rover can be seen trundling across the pockmarked surface of Earth's satellite in pictures released today by the Chinese Academy of Sciences. The successful soft-landing of the Chang'e 3 lander in December made China the third country to complete a lunar rover mission after the United States and Russia. Chasing the horizon: The sky over the Moon appears inky black in this picture showing the Jade Rabbit rover trundling across the lunar surface taken from the Chang'e 3 lander and released today by Chinese scientists.
Keyword: China ### Relevant Fact:
\end{spverbatim}
\vspace{10px}

\noindent \textbf{Generated Output:}
\begin{spverbatim}
Knock off or not, there's no denying the ambition of the China's Moon mission, which on December 14 achieved the first lunar soft landing in nearly four decades and aims at a manned mission by 2025.
\end{spverbatim}

\subsubsection{RTE}
\textbf{Input Prompt:}
\begin{spverbatim}
Input: No Weapons of Mass Destruction Found in Iraq Yet.
Output: Weapons of Mass Destruction Found in Iraq.

Input: CBS newsman Harry Reasoner is returning to his Iowa hometown to get married Saturday.
Output: In other words,
\end{spverbatim}
\vspace{10px}

\noindent \textbf{Generated Output:}
\begin{spverbatim}
Harry Reasoner is getting married.
\end{spverbatim}

\subsubsection{WiC}
\textbf{Input Prompt:}
\begin{spverbatim}
Word: meet
Example: The company agrees to meet the cost of any repairs.

Word: skin
Definition 1:
\end{spverbatim}
\vspace{10px}

\noindent \textbf{Generated Output:}
\begin{spverbatim}
the outer covering of the body of a vertebrate animal.
Example: The skin of a snake.
\end{spverbatim}

\subsubsection{WSC}
\textbf{Input Prompt:}
\begin{spverbatim}
Noun: Sam ### Pronoun: he ### Sample Sentence: Sam took French classes from Adam , because he was eager to speak it fluently.

Noun: Ann ### Pronoun: She ###  Sentence where She refers to Ann:
\end{spverbatim}
\vspace{10px}

\noindent \textbf{Generated Output:}
\begin{spverbatim}
It seems that Ann has had a run of bad luck. She never wins at cards.
\end{spverbatim}
\end{spverbatim}
\end{spverbatim}
\end{spverbatim}
\end{spverbatim}
\end{spverbatim}
\end{spverbatim}
\end{spverbatim}
\end{spverbatim}
\end{spverbatim}
\end{spverbatim}
\end{spverbatim}
\end{spverbatim}
\end{spverbatim}
\end{spverbatim}
\end{spverbatim}
\end{spverbatim}
\end{spverbatim}
\end{spverbatim}
\end{spverbatim}
\end{spverbatim}
\end{spverbatim}
\end{spverbatim}
\end{spverbatim}
\end{spverbatim}
\end{spverbatim}
\end{spverbatim}
\end{spverbatim}
\end{spverbatim}
\end{document}